\pdfoutput=1
\documentclass[11pt,a4paper]{article}
\usepackage{naaclhlt2019}
\usepackage{natbib}
\usepackage{times}
\usepackage{latexsym}
\usepackage{comment}
\usepackage{color}
\usepackage{todonotes}
\usepackage{multirow}

\usepackage{graphicx}
\usepackage{tabularx}
\usepackage{arydshln}

\usepackage{url}

\aclfinalcopy 
\def\aclpaperid{69} 

\title{Word Usage Similarity Estimation \\ 
with Sentence Representations and Automatic Substitutes}

\author{Aina Gar\'i Soler$^{1}$, Marianna Apidianaki$^{1,2}$ and Alexandre Allauzen$^{1}$ \\
\\
$^{1}$LIMSI, CNRS, Univ. Paris Sud, Universit\'e Paris-Saclay, F-91405 Orsay, France\\
$^{2}$LLF, CNRS, Univ. Paris Diderot\\
\texttt{\{aina.gari,marianna,allauzen\}@limsi.fr}}  
  
\date{}

\begin{document}
\maketitle
\begin{abstract}
Usage similarity estimation addresses the semantic proximity of word instances in different contexts. We apply contextualized (ELMo and BERT) word and sentence embeddings to this task, and propose supervised models that leverage these representations for prediction. 
Our models are further assisted by lexical substitute annotations automatically assigned to word instances by context2vec, a neural model that relies on a bidirectional LSTM. We perform an extensive comparison of existing word and sentence representations on benchmark datasets addressing both graded and binary similarity. The best performing models outperform previous methods in both settings.
\end{abstract}

\section{Introduction}

Traditional word embeddings, like Word2Vec and GloVe, merge different meanings of a word in a single vector representation~\citep{mikolov2013efficient,pennington2014glove}. These pre-trained embeddings are fixed, and stay the same independently of the context of use. Current contextualized sense representations, like ELMo and BERT, go to the other extreme and model meaning as word usage~\citep{peters2018deep,devlin2018bert}. They provide a dynamic representation of word meaning adapted to every new context of use.

\begin{figure}[bt]
\includegraphics[width=7.6cm]{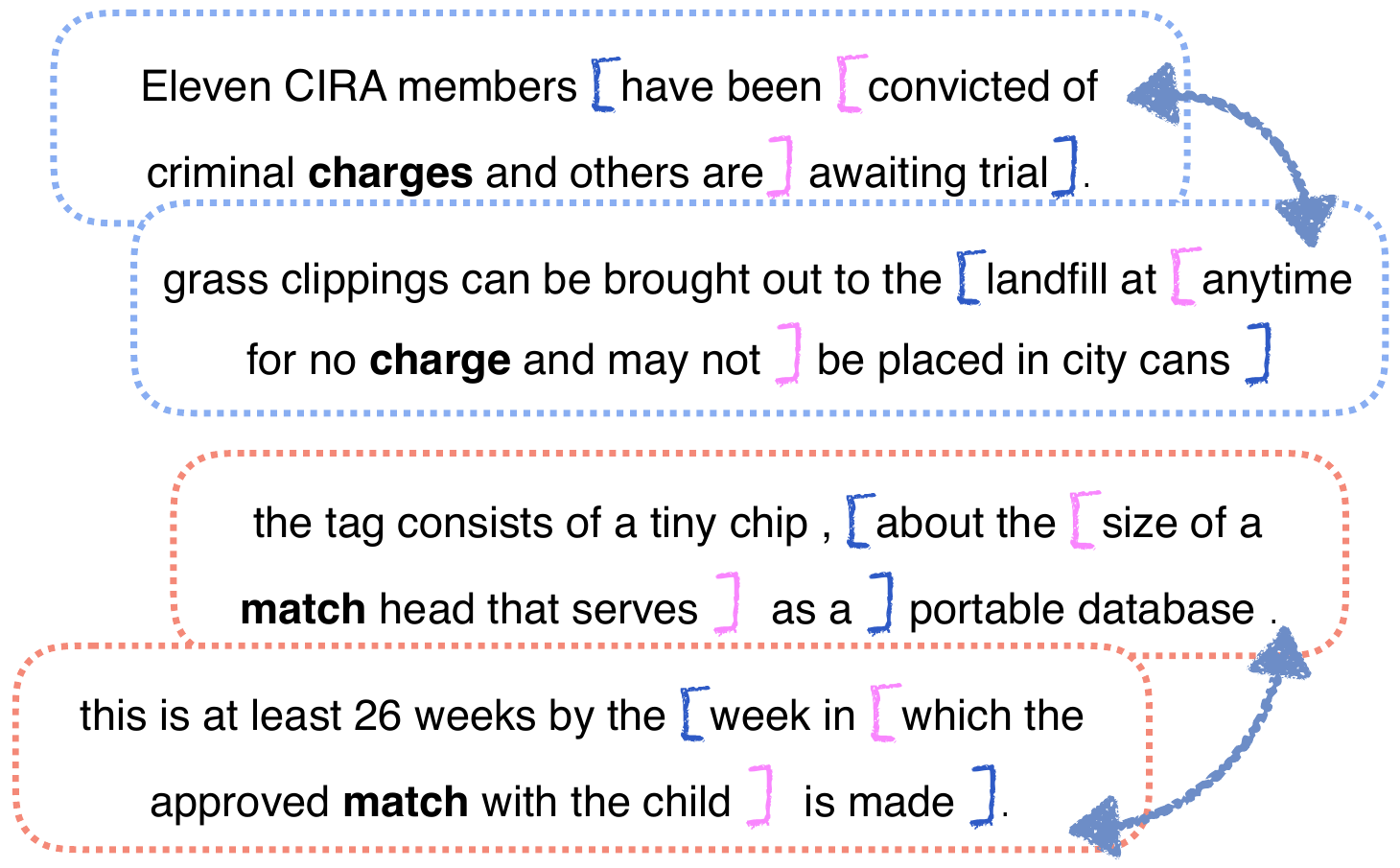}
\caption{We use contextualized word representations built from the whole sentence or smaller windows around the target word for usage similarity estimation, combined with  automatic substitute annotations.} \label{tab:example_cw}
\end{figure}

In this work, we perform an extensive comparison of existing static and dynamic embedding-based meaning representation methods on the usage similarity (Usim) task, which involves estimating the semantic proximity of word instances in different contexts \cite{erketal2009}. Usim differs from a classical Semantic Textual Similarity task \cite{agirreetal2016} by the focus on a particular word in the sentence.  
We evaluate on this task word and context representations obtained using pre-trained uncontextualized word embeddings (GloVe)  \cite{pennington2014glove}, with and without dimensionality reduction (SIF) \citep{arora2017asimple}; context representations obtained from a bidirectional LSTM (context2vec) \cite{melamud2016context2vec}; contextualized word embeddings derived from a LSTM bidirectional language model (ELMo) \cite{peters2018deep} and generated by a Transformer (BERT) \cite{devlin2018bert}; doc2vec \citep{le2014distributed} and  Universal Sentence Encoder representations \citep{cer2018universal}. 
All these embedding-based methods provide direct assessments of usage similarity. 
The best representations are used as features in  supervised models for Usim prediction, trained on similarity judgments.

We combine direct Usim assessments, made by the embedding-based methods, with a substitute-based Usim approach. Building up on previous work that used manually selected in-context substitutes as a proxy for Usim  \cite{J13-3003,J16-2003}, we propose to automatize the annotation collection step in order to scale up the method and make it operational on unrestricted text. We exploit annotations assigned to words in context by the context2vec lexical substitution model, which relies on word and context representations learned by a  bidirectional LSTM from a large corpus \citep{melamud2016context2vec}. 
 
The main contributions of this paper can be summarized as follows: 
\begin{itemize}
\setlength\itemsep{0.05em}
    \item we provide a direct comparison of a wide range of word and sentence representation methods on the Usage Similarity (Usim) task and show that current contextualized representations can successfully predict Usim;
    \item we propose to automatize, and scale up, previous substitute-based Usim prediction methods;
    \item we propose supervised models for Usim prediction which integrate embedding and lexical substitution features;
    \item we propose a methodology for  collecting new training data for supervised Usim prediction from datasets annotated for related tasks. 
\end{itemize}

We test our models on benchmark  datasets containing gold graded and binary word Usim judgments  \cite{J13-3003,PilehvarandCamachoCollados}. From the compared embedding-based approaches, the BERT model gives best results on both types of data, providing a straightforward way for word usage similarity calculation. Our supervised model performs on par with BERT on the graded and binary Usim tasks, 
when using embedding-based representations and clean lexical substitutes.

\section{Related Work}

Usage similarity is a means for representing word meaning which involves assessing in-context semantic similarity, rather than mapping to word senses from external inventories \cite{erketal2009,J13-3003}. This methodology followed from the gradual shift from word sense disambiguation models that would select the best sense in context from a dictionary, to models that reason about meaning by solely relying on distributional similarity \cite{D08-1094,mitchellLapata}, or allow multiple sense interpretations \cite{jurgens2014analysis}. In \newcite{erketal2009}, the idea is to model meaning in context in a way that captures different degrees of similarity to a word sense, or between word instances. 

Due to its high reliance on context, Usim can be viewed as a semantic textual similarity (STS) \cite{agirreetal2016} task with a focus on a specific word instance. This connection motivated us to apply methods initially proposed for sentence similarity to Usim prediction. 
More precisely, we build sentence representations using different types of word and sentence embeddings, ranging from the classical word-averaging approach with traditional word embeddings \cite{pennington2014glove}, to more recent contextualized word representations \citep{peters2018deep,devlin2018bert}.
We explore the contribution of each separate method for Usim prediction, and use the best performing ones as features in supervised models. These are trained on sentence pairs labelled with Usim judgments  \cite{erketal2009} to predict the similarity of new word instances. 

Previous attempts to automatic Usim prediction involved obtaining vectors encoding a distribution of topics for every target word in context \cite{U12-1006}. In this work, Usim was approximated by the cosine similarity of the resulting topic vectors. We show how contextualized representations, and the supervised model that uses them as features, outperform topic-based methods on the graded Usim task. 

We combine the embedding-based direct Usim assessment methods with substitute-based representations obtained using an unsupervised lexical substitution model. \newcite{J16-2003} showed it is possible to model usage similarity using manual substitute annotations for words in context.
In this setting, the set of substitutes proposed for a word instance describe its specific meaning, while similarity of substitute annotations for different instances points to their semantic  proximity.\footnote{McCarthy et al. use the substitute annotations as features for predicting Usim, clustering instances  and estimating the partitionability of words into senses. This offers a way to distinguish between lemmas with distinct senses and others with fuzzy semantics, which would be more challenging in annotation tasks and automatic processing.} 
We follow up on this work and propose a way to use substitutes for Usim prediction on unrestricted text, bypassing the need for manual annotations.
Our method relies on substitute annotations proposed by the context2vec model \citep{melamud2016context2vec}, which uses word and context representations learned by a bidirectional LSTM from a large corpus (UkWac) \citet{baroni2009wacky}.

\begin{table*}[t!] 
\centering
\scalebox{0.9}{
\begin{tabular}{m{5.9cm} m{10.9cm}}
   \begin{center}Sentences\end{center} & \begin{center}Substitutes\end{center}\\
  \hline
The local {\bf papers} took photographs of the footprint.
& {\sc gold}: newspaper, journal \newline
{\sc auto-lscnc}: press, newspaper, news, report, picture 
\newline
{\sc auto-ppdb}: newspaper, newsprint
\vspace{1mm}
\\
 \cline{1-2}
Now Ari Fleischer, in a pitiful letter to the {\bf paper}, tries to cast Milbank as the one getting his facts wrong.
& {\sc gold}: newspaper, publication  \newline 
{\sc auto-lscnc}: press, newspaper, news, article, journal, thesis, periodical, manuscript, document  
\newline 
{\sc auto-ppdb}: newspaper 
\vspace{1mm}
\\ 
\hline
\hline

This is also at the very essence or heart of being a {\bf coach}. &
{\sc gold}: trainer, tutor, teacher  \newline
{\sc auto-lscnc}: teacher, counsellor, trainer, tutor, instructor 
\newline
{\sc auto-ppdb}: trainer, teacher, mentor, coaching 
\vspace{1mm}
\\
\cline{1-2}
We hopped back onto the {\bf coach} -- now for the boulangerie! & 
{\sc gold}: coach, bus, carriage \newline
{\sc auto-lscnc}: bus, car, carriage, transport  newline 
{\sc auto-ppdb}: bus, train, wagon, lorry, car, truck, carriage, vehicle  \\
\end{tabular}}
\caption{Example pairs of highly similar and dissimilar usages from the Usim dataset \citep{J13-3003} for the nouns {\it paper} (Usim score $=4.34$) and \textit{coach.n} (Usim score $=1.5$), with the substitutes  assigned by the annotators ({\sc gold}). For comparison, we give the  substitutes selected for these instances by the automatic substitution method (context2vec) used in our experiments from two different pools of substitutes ({\sc auto-lscnc} and {\sc ppdb}). More details on the automatic substitution configurations are given in Section \ref{subusim}. \label{tab:exampleusimsubs} } 

\end{table*}

\section{Data}

\subsection{The LexSub and Usim Datasets}
\label{lexsubdataset}

We use the training and test datasets of the SemEval-2007 Lexical Substitution (LexSub) task \cite{S07-1009}, which contain instances of target words in sentential context hand-labelled with meaning-preserving substitutes. 
A subset of the LexSub data (10 instances x 56 lemmas) has additionally been annotated with graded pairwise Usim judgments \cite{J13-3003}. Each sentence pair received a rating (on a scale of 1-5) by multiple annotators, and the average judgment for each pair was retained. \newcite{J16-2003} derive two additional scores from  Usim annotations that denote how easy it is to partition a lemma's usages into sets describing distinct senses: {\bf Uiaa}, the inter-annotator agreement for a given lemma, taken as the average pairwise Spearman's $\rho$ correlation between ranked judgments of the annotators; and {\bf Umid}, the proportion of mid-range judgments over all instances for a lemma and all annotators. 

In our experiments, we use 2,466 sentence pairs from the Usim data for training, development and testing of different automatic Usim prediction methods. 
Our models rely on substitutes automatically assigned to words in context using context2vec \citep{melamud2016context2vec}, and on various word and sentence embedding representations. We also train a model using the gold substitutes, to test how well our models perform when substitute quality is high.  
Performance of the different models is evaluated by measuring how well they approximate the Usim scores assigned by annotators. Table \ref{tab:exampleusimsubs} shows examples of sentence pairs from the Usim dataset \cite{J13-3003} with the {\sc gold} substitutes and Usim scores assigned by the annotators. 
The Usim score is high for  similar instances, and decreases for instances that describe different meanings. The  semantic proximity of two instances is also reflected in the similarity of their substitutes sets. 
For comparison, we also give in the Table the substitutes selected for these instances by the automatic context2vec substitution method used in our experiments (more details in Section \ref{subusim}). 

\subsection{The Concepts in Context Corpus}
\label{coincocorpus}

Given the small size of the Usim dataset, we extract additional training data for our models from the 
Concepts in Context (CoInCo) corpus  \citep{kremer2014substitutes}, a subset of the \textsc{MASC} corpus \citep{ide2008masc}. CoInCo contains manually selected substitutes for all content words in a sentence, but provides no usage similarity scores that could be used for training. We construct our supplementary training data as follows: we gather all instances of a target word in the corpus with at least four substitutes, and keep pairs with (1) no overlap in substitutes, and (2) minimum 75\% substitute overlap.\footnote{Full overlap is rare since annotators propose somewhat different sets of substitutes, 
even for instances with the same meaning. Full overlap is observed for only 437 of all considered CoInCo pairs (0.3\%).} We view the first set of pairs as examples of completely different usages of a word ({\sc diff}), and the second set as examples of identical usages ({\sc same}). 
The two sets are unbalanced in terms of number of instance pairs (19,060 vs. 2,556). We balance them by keeping in {\sc diff} the 2,556 pairs with the highest number of substitutes. 

We also annotate the data with substitutes using context2vec \citep{melamud2016context2vec}, as described in Section \ref{subusim}.
We apply an additional filtering to the sentence pairs extracted from CoInCo, discarding instances of words that are not in the context2vec vocabulary and have no embeddings. We are left with 2,513 pairs in each class (5,026 in total). We use 80\% of these pairs (4,020) together with the Usim data to train our supervised Usim models described in Section \ref{supervisedusim}.\footnote{We will make the dataset available at \url{https://github.com/ainagari}. 20\% of the extracted examples were kept aside for development and testing purposes.}

\subsection{The Word-in-Context dataset}
\label{wicdataset}

The third dataset we use in our experiments is the recently released Word-in-Context (WiC) dataset \cite{PilehvarandCamachoCollados}, version 0.1. WiC provides pairs of contextualized target word instances describing the same or different meaning, framing in-context sense identification as a binary classification task. For example, a sentence pair for the noun {\it stream} is: [`{\underline {Stream}} of consciousness' -- `Two {\underline {streams}} of development run through American history']. 
A system is expected to be able to identify that {\it stream} does not have the same meaning in the two sentences. 

WiC sentences were extracted from example usages in WordNet \cite{Fellbaum1998}, VerbNet \cite{KipperSchuler2006}, and Wiktionary. Instance pairs were automatically labeled as positive (T) or negative (F) (corresponding to the same/different sense) using information in the lexicographic resources, such as presence in the same or different synsets. 
Each word is represented by at most three instances in WiC, and repeated sentences are excluded. 
It is important to note that meanings  represented in the WiC dataset are coarser-grained than WordNet senses. This was ensured by excluding WordNet synsets describing highly similar meanings (sister senses, and senses belonging to the same supersense). The human-level performance upper-bound on this binary task, as measured on two 100-sentence samples, is 80.5\%. Inter-annotator agreement is also high, at 79\%.
The dataset comes with an official train/dev/test split containing 7,618, 702 and 1,366 sentence pairs, respectively.\footnote{The test portion of WiC had not been released at the time of submission. We contacted the authors and ran the evaluation on the official test set, to be able to compare to results reported in their paper \cite{PilehvarandCamachoCollados}.}

\section{Methodology}

We experiment with two ways of predicting usage similarity: an unsupervised approach which relies on the cosine similarity of different kinds of word and sentence 
representations, and provides direct Usim assessments; and supervised models that combine embedding similarity with features based on substitute overlap. 
We present the direct Usim prediction methods in Section \ref{dirUsim}. In Section \ref{subusim}, we describe how substitute-based features were extracted, and in Section \ref{supervisedusim}, we introduce the supervised Usim models.

\subsection{Direct Usage Similarity  Prediction}
\label{dirUsim}

In the unsupervised Usim prediction setting, we apply different types of pre-trained word and sentence embeddings as follows: we compute an embedding for every sentence in the Usim dataset,
and calculate the pairwise cosine similarity between the sentences available for a target word. 
Then, for every embedding type, we  measure the correlation between 
sentence similarities and gold usage similarity  judgments in the Usim dataset, using Spearman's \begin{math} \rho \end{math} correlation coefficient. We experiment with the following embedding types. 

\vspace{2mm}
\noindent {\bf GloVe} embeddings are uncontextualized word representations which merge all senses of a word in one vector \citep{pennington2014glove}. We use 300-dimensional GloVe embeddings pre-trained on Common Crawl (840B tokens).\footnote{\url{https://nlp.stanford.edu/projects/glove/}} The representation of a sentence is obtained by averaging the GloVe embeddings of the words in the sentence.

\vspace{2mm}
\noindent \textbf{SIF} (Smooth Inverse Frequency) embeddings are sentence representations built by applying dimensionality reduction to a weighted average of uncontextualized embeddings of words in a sentence \citep{arora2017asimple}. We use SIF in combination with GloVe vectors.

\vspace{2mm}
\noindent \textbf{Context2vec}  embeddings \citep{melamud2016context2vec}. The context2vec model learns embeddings for words and their sentential contexts simultaneously. The resulting representations reflect: a) the similarity between potential fillers of a sentence with a blank slot, 
and b) the similarity of contexts that can be filled with the same word. We use a context2vec model pre-trained on the UkWac corpus \citep{baroni2009wacky} \footnote{\url{http://u.cs.biu.ac.il/~nlp/resources/downloads/context2vec/}} to compute embeddings for sentences with a blank at the target word's position. 

\vspace{2mm}
\noindent \textbf{ELMo} (Embeddings from Language Models) representations are contextualized word embeddings derived from the internal states of an LSTM bidirectional language model (biLM) \citep{peters2018deep}. In our experiments, we use a pre-trained 512-dimensional 
biLM.\footnote{\url{https://allennlp.org/elmo}} Typically, the best linear combination of the layer representations for a word is learned for each end task in a supervised manner. 
Here, we use out-of-the-box embeddings (without tuning) and experiment with the top layer, and with the average of the three hidden layers.
We represent a sentence in two ways: by the contextualized ELMo embedding obtained for the target word, and by the average of ELMo embeddings for all words in a sentence. 

\vspace{2mm}
\noindent \textbf{BERT} (Bidirectional Encoder Representations from Transformers) \citep{devlin2018bert}. BERT representations are generated by a 12-layer bidirectional Transformer encoder that jointly conditions on both left and right context in all layers.\footnote{This is an important difference with the ELMo architecture which concatenates a left-to-right and right-to-left model.} BERT can be fine-tuned to specific end tasks, or its contextualized word representations can be used directly in applications, similar to ELMo. We try different layer combinations and create sentence representations, in the same way as for ELMo: using either the BERT embedding of the target word, or the average of the BERT embeddings for all words in a sentence.

\vspace{2mm}
\noindent \textbf{Universal Sentence Encoder (USE)} 
makes use of a Deep Averaging Network (DAN) encoder trained to create sentence representations by means of multi-task learning \citep{cer2018universal}. USE has been shown to improve performance on different NLP tasks using transfer learning.\footnote{\url{https://tfhub.dev/google/universal-sentence-encoder/2}} 

\vspace{2mm}
\noindent \textbf{doc2vec} is an extension of word2vec to the sentence, paragraph or document level \citep{le2014distributed}. One of its forms, {\it dbow} (distributed bag of words), is based on the skip-gram model, where it adds a new feature vector representing a document. We use a dbow model trained on English Wikipedia released by \citet{lau2016empirical}.\footnote{\url{https://github.com/jhlau/doc2vec}}

\vspace{1mm}

We test the above models with representations built from the whole sentence, and using a smaller context window (cw) around the target word. Sentences in the WiC dataset are quite short (7.9 $\pm$ 3.9 words), but the length of sentences in the Usim and CoInCo datasets varies a lot (27.4 $\pm$ 13.2 and 18.8 $\pm$ 10.2, respectively). 
We want to check whether information surrounding the target word in the sentence is more relevant, and sufficient for Usim estimation. 
We focus on the words in a context window of $\pm$ 2, 3, 4 or 5 words at each side of a target word. Then, we collect their word embeddings to be averaged (for GloVe, ELMo and BERT), or  derive an embedding from this specific window instead of the whole sentence (for USE). 

We approximate Usim by measuring the cosine similarity of the resulting context representations. We compare the performance of these direct assessment methods on the Usim dataset and report the results in Section \ref{evalsec}.

\subsection{Substitute-based Feature Extraction} \label{subusim}

Following up on McCarthy et al.'s \shortcite{J16-2003} sense clusterability work, we also experiment with a substitute-based approach for Usim prediction. McCarthy et al. showed that manually selected substitutes for word instances in context can be used as a proxy for Usim. Here, we propose an approach to obtain these annotations automatically that can be applied to the whole vocabulary.

\vspace{2mm}
\noindent {\bf Automatic LexSub} We generate rankings of candidate substitutes for words in context using the context2vec method \cite{melamud2016context2vec}. 
The original method selects and ranks substitutes from the whole vocabulary. To facilitate comparison and evaluation, 
we use the following pools of candidates:
(a) all substitutes that were proposed for a word in the LexSub and CoInCo annotations (we call this substitute pool {\sc auto-lscnc}); (b) the paraphrases of the word in the Paraphrase Database (PPDB) XXL package \cite{N13-1092,P15-2070} ({\sc auto-ppdb}).\footnote{
\url{http://paraphrase.org/}} 
In the WiC experiments, where no substitute annotations are available, we only use PPDB paraphrases ({\sc auto-ppdb}). 
We obtain a context2vec embedding for a sentence by replacing the target word with a blank.
{\sc auto-lscnc} substitutes are high-quality since they were extracted from the manual LexSub and CoInCo annotations. They are semantically similar to the target, and context2vec just needs to rank them according to how well they fit the new context. 
This is done by measuring the cosine similarity between each substitute's context2vec word embedding and the context embedding obtained for the sentence.

The {\sc auto-ppdb} pool contains paraphrases from PPDB XXL, which were automatically extracted from parallel corpora \cite{N13-1092}. Hence, this pool contains noisy paraphrases that should be ranked lower. To this end, we use in this setting the original context2vec scoring formula which also accounts for the similarity between the target word and the substitute:

\begin{equation} \label{c2v_formula}
c2v score = \frac{cos(s,t) + 1}{2} \times \frac{cos(s,C) + 1}{2}
\end{equation}

\noindent In formula (1), $s$ and $t$ are the word embeddings of a substitute and the target word, and $C$ is the context2vec vector of the context. 
Following this procedure, context2vec produces a ranking of candidate substitutes for each target word instance in the Usim, CoInCo and WiC datasets, according to their fit in context. 
Every candidate is assigned a score, with substitutes that are a good fit in a specific context being higher-ranked than others. 
For every new target word instance, context2vec ranks {\em all} candidate substitutes available for the target  in each pool. Consequently, the automatic annotations produced for different instances of the target  include the same set of substitutes, but in different order. This does not allow for the use of measures based on substitute overlap, which were shown to be useful for Usim prediction in \newcite{J16-2003}. In order to  use this type of measures, we propose ways to filter the automatically generated rankings, and keep for each instance only substitutes that are a good fit in context.

\vspace{2mm}
\noindent {\bf Substitute Filtering} We test different filters to discard low quality substitutes from the annotations proposed by context2vec for each instance.

\begin{itemize}
\setlength\itemsep{1pt}
\item {\bf PPDB 2.0 score}: Given a 
ranking $R$ of $n$ substitutes $R = [s_1, s_2, ..., s_n]$ proposed by context2vec, we form pairs of substitutes in adjacent positions \{${s_i} \leftrightarrow  {s_{i+1}}$\}, and check  whether they exist as paraphrase pairs in PPDB. 
We expect substitutes that are paraphrases of each other to be similarly ranked.
If $s_i$ and $s_{i+1}$ are not paraphrases in PPDB, we keep all substitutes up to $s_i$ and use this as a cut-off point, discarding substitutes present from position $s_{i+1}$ onwards in the ranking.

\item \textbf{GloVe word embeddings}: We measure the cosine similarity ({\it cosSim}) between GloVe embeddings of adjacent substitutes \{${s_i} \leftrightarrow{s_{i+1}}$\} in the ranking $R$ obtained for a new instance.
We first compare the similarity of the first pair of substitutes ({\it cosSim}({$s_1,s_2$})) to a lower bound similarity threshold T. If {\it cosSim}({$s_1,s_2$}) exceeds T, we assume that $s_1$ and $s_2$ have the same meaning, and use {\it cosSim}({$s_1,s_2$}) as a   reference similarity value, $S$, for this instance.
The middle point between the two values, $M = (T+S)/2$, is then used as a threshold to determine whether there is a shift in meaning in subsequent pairs. 
If $cosSim(s_{i},s_{i+1}) < M$, for $i > 1$, then only the higher ranked substitute ($s_{i}$) is retained and all subsequent substitutes in the ranking are discarded. The intuition behind this calculation is that if $cosSim$ is much lower than the reference $S$ (even if it exceeds $T$), substitutes possibly have different senses. 

\item {\bf Context2vec score}: This filter uses the score assigned by context2vec to each substitute, reflecting how good a fit it is in each context. context2vec scores vary a lot across instances, it is thus not straightforward to choose a threshold. We instead refer to the scores assigned 
to adjacent pairs of substitutes in the ranking produced for each instance, $R = [s_1, s_2, ..., s_n]$. We view the pair with the biggest difference in scores as the cut-off point, considering it reflects a degradation in substitute fit.
We retain only substitutes up to this point.

\item {\bf Highest-ranked $X$ substitutes}. We also test two simple baselines, which consist in keeping the 5 and 10 highest-ranked substitutes  for each instance. 

\end{itemize}

We test the efficiency of each filter on the portion of the LexSub dataset \cite{S07-1009} that was not annotated for Usim. 
We compare the substitutes retained for each instance after filtering to its gold LexSub susbtitutes using the F1-score, and the proportion of false positives out of all positives. Filtering results are reported in Appendix \ref{app:filtering}. 
The best filters were GloVe word embeddings ($T=0.2$) for {\sc auto-lscnc}, and the PPDB filter for {\sc auto-ppdb}.

\vspace{2mm}
\noindent {\bf Feature Extraction} After annotating the Usim sentences with context2vec and filtering, we extract, for each sentence pair ($S_1$, $S_2$), a set of features related to the amount of substitute overlap. 

\begin{itemize}
\setlength{\parskip}{0pt}
\item \textbf{Common substitutes}. The proportion of shared substitutes between  two sentences. 
\item \textbf{GAP score}. The average of the Generalized Average Precision (GAP) score \citep{kishida2005property} taken in both directions ($GAP(S_1, S_2)$ and $GAP(S_2, S_1)$). GAP is a measure that compares two rankings considering not only the order of the ranked elements but also their weights. It ranges from 0 to 1, where 0 means that rankings are completely different and 1 indicates perfect agreement. 
We use the frequency in the manual Usim annotations (i.e. the number of annotators who proposed each substitute) as the weight for gold substitutes, and the context2vec score for automatic substitutes. We use the GAP implementation from \citet{melamud2015simple}.

\item \textbf{Substitute cosine similarity}. We form substitute pairs ($S_1$ $\leftrightarrow$ $S_2$)  and calculate the average of their GloVe cosine similarities.
This feature shows the semantic similarity of substitutes, even when overlap is low. 
\end{itemize}

\subsection{Supervised Usim Prediction}
\label{supervisedusim}

We train linear regression models to predict Usim scores for word instances in different contexts using as features the cosine similarity of the different representations  in Section \ref{dirUsim}, and the substitute-based features in \ref{subusim}. 
For training, we use the Usim dataset on its own (cf. Section \ref{lexsubdataset}), and combined with the additional training examples extracted from CoInCo (cf. Section \ref{coincocorpus}). 

To be able to evaluate the performance of our models separately for each of the 56 target words in the Usim dataset, we train a separate model for each word in a leave-one-out setting. Each time, we use 2,196 pairs for training, 225 for development and 45 for testing.\footnote{With the exception of 4 lemmas which had 36 pairs, and one which had 44.} Each model is evaluated on the sentences corresponding to the left out target word. We report results of these experiments in Section \ref{evalsec}. The performance of the model with context2vec substitutes from the two substitute pools is compared to that of the model with gold substitute annotations. We replicate the experiments by adding CoInCo data to the Usim training data.

To test the contribution of each feature, we perform an ablation study on the 225 Usim sentence pairs of the development set, which cover the full spectrum of Usim scores (from 1 to 5). We report results of the  feature ablation in Appendix \ref{app:ablation}.

We also build a model for the binary Usim task on the WiC dataset \cite{PilehvarandCamachoCollados}, using the official train/dev/test split. 
We train a logistic regression classifier on the training set, and use the development set to select the best among several feature combinations. We report results of the best performing models on the WiC test set in Section \ref{evalsec}.
For instances in WiC where no PPDB substitutes are available (133 out of 1,366 in the test set) we back off to a model that only relies on the embedding features.

\begin{table}[t] 
\centering
\scalebox{0.9}{
\begin{tabular}{ll|c} 
 Context & Embeddings & Correlation \\
\hline
  \multirow{7}{*}{Full sentence} & GloVe & 0.142   \\
   & SIF & 0.274  \\
  & c2v  & 0.290 \\
   & USE & 0.272  \\
   & doc2vec & 0.124 \\
   & ELMo av & 0.254 \\
   & BERT av 4& 0.289\\
   \hline
   \multirow{3}{*}{Target word}& ELMo av  & 0.166\\
   & ELMo top & 0.177 \\
   & BERT top & 0.514 \\
& BERT av 4 & \textbf{0.518}\\
   \hline
   \multirow{1}{*}{cw=2} & ELMo top & 0.289\\
   \hline
   \multirow{3}{*}{cw=3 (incl. target)} & GloVe & 0.180  \\
   & ELMo av & 0.280 \\
   & BERT av 4 & 0.395 \\
\hline
   \multirow{4}{*}{cw=5 (incl. target)} & USE &  0.221 \\
   & ELMo av & 0.266\\
   & ELMo top & 0.263 \\
   & BERT top & 0.309 \\
   \hline
\end{tabular}}
\caption{Spearman $\rho$ correlation of different sentence and word embeddings on the Usim dataset using different context window sizes (cw). For BERT and ELMo, \textit{top} refers to the top layer, and \textit{av} refers to the average of layers (3 for ELMo, and the last 4 for BERT).\label{tab:embed_correlations}} 
\end{table}

\begin{table*}[bt!]
\centering
\scalebox{0.9}{
\begin{tabular}{ l l ||c|c|c} 
\multirow{2}{*}{Training set} & \multirow{2}{*}{Features} & \multirow{2}{*}{Gold} & c2v & c2v \\
   &  &  & {\sc auto-lscnc} &  {\sc auto-ppdb} \\
  \hline
  \multirow{3}{*}{Usim} & Substitute-based & 0.563 & 0.273 & 0.148\\
   & Embedding-based & 0.494 & 0.494 & 0.494\\
& Combined & \textbf{0.626} & 0.501 & 0.493 \\
   \hline
   \multirow{3}{*}{Usim + CoInCo} & Substitute-based & - & 0.262 & 0.129\\
   & Embedding-based & - & 0.495 & 0.495\\
   & Combined & - & 0.501 & 0.491\\

\end{tabular}
}
\caption{{\bf Graded Usim results}: Spearman's $\rho$ correlation results between supervised model predictions and graded  annotations on the Usim test set. The first column reports results obtained using gold substitute annotations for each target word instance. The last two columns give results with automatic substitutes selected among all substitutes proposed for the word in the LexSub and CoInCo datasets ({\sc auto-lscnc}), or paraphrases in the PPDB XXL package ({\sc auto-ppdb}). The Embedding-based configuration uses cosine similarities from BERT and context2vec.}  
\label{tab:sup_results_usim}
\end{table*}

\begin{table*}[ht!]
\centering
\scalebox{0.9}{
\begin{tabular}{ l l |  c}
  Training set & Features  &  Accuracy\\ 
  \hline
  \multirow{4}{*}{WiC}   & Embedding-based & 63.62 \\
   & Combined & \textbf{64.86} \\
   & DeConf embeddings 
   \citep{PilehvarandCamachoCollados} & 59.4   \\
   & Random baseline \citep{PilehvarandCamachoCollados} &  50.0 \\
   
   \hline
   
   \multirow{2}{*}{WiC + CoInCo}  & Embedding-based & 63.69 \\
   & Combined & 64.42 \\

\end{tabular}
}
\caption{{\bf Binary Usim results}: Accuracy of models on the WiC test set. The Embedding-based configuration includes cosine similarities of BERT target and USE. The Combined setting uses, in addition, substitute overlap features ({\sc auto-ppdb}).}
\label{tab:sup_results_wic}
\end{table*}

\section{Evaluation} \label{evalsec}

\noindent{\bf Direct Usim Prediction} 

Correlation results between Usim judgments and the cosine similarity of the embedding representations described in Section \ref{dirUsim} are found in Table \ref{tab:embed_correlations}. Detailed results for all context window combinations are given in Appendix \ref{app:direct}.
We observe that target word BERT embeddings give best performance in this task.  
Selecting a context window around (or including) the target word does not always help, on the contrary it can harm the models.
Context2vec sentence representations are the next best performing representation, after BERT, but their correlation is much lower. 
The simple GloVe-based SIF approach for sentence representation, which consists in applying dimensionality reduction to a weighted average of GloVe vectors of the words in a sentence, is much superior to the simple average of GloVe vectors and even better than doc2vec sentence representations, obtaining a correlation comparable to that of USE.

\vspace{2mm}
\noindent{\bf Graded Usim} To evaluate the performance of our supervised models, we measure the correlation of the  predictions with human similarity judgments on the Usim dataset using Spearman's $\rho$. 
Results reported 
in Table \ref{tab:sup_results_usim} are the average of the correlations obtained for each target word with gold and automatic substitutes (from the two  substitute pools), and for each type of features, substitute-based and embedding-based (cosine similarities from BERT and context2vec).
We also report  results with the additional CoInCo training data. 
Unsurprisingly, the best results are obtained by the methods that use the gold substitutes.
This is consistent with previous analyses by \citet{erketal2009} who found  overlap in manually-proposed substitutes to correlate with Usim judgments.
The lower performance of features that rely on automatically selected substitutes ({\sc auto-lscnc} and {\sc auto-ppdb}) demonstrates the impact of substitute quality on the contribution of this type of features. 
The addition of CoInCo data does not seem to help the models, as results are slightly lower than in the only Usim setting. This can be due to the fact that CoInCo data contains only extreme cases of similarity ({\sc same/diff}) and no intermediate ratings. 
The slight improvement in the combined settings over embedding-based models is not significant in {\sc auto-lscnc} substitutes, but it is for gold substitutes (p $<$ 0.001).\footnote{As determined by paired t-tests, after verifying the normality of the differences with the Shapiro-Wilk test}

For comparison to the topic-modelling approach of \citet{U12-1006}, we evaluate on the 34 lemmas used in their experiments. They report a correlation calculated over all instances. 
 With the exception of the substitute-only setting with PPDB candidates, all of our Usim models get higher correlation than their model ($\rho=0.202$),  with $\rho=0.512$ for the combination of {\sc auto-lscnc} substitutes and embeddings. The average of the per target word correlation in \citet{U12-1006}  ($\rho=0.388$) is still lower than that of our {\sc auto-lscnc} model in the combined setting ($\rho=0.500$).

\vspace{2mm}
\noindent {\bf Binary Usim} We evaluate the predictions of our binary classifiers by measuring  accuracy on the test portion of the WiC dataset. 
Results for the best  configurations for each training set are reported in Table \ref{tab:sup_results_wic}. Experiments on the development set showed that target word BERT representations and USE sentence embeddings are the best-suited for WiC.
Therefore, `embedding-based features' here refers to these two representations. Results on the development set can be found in Appendix \ref{app:devwic}. 
All configurations obtain higher accuracy than the previous best reported result on this dataset (59.4) \citep{PilehvarandCamachoCollados}, obtained using DeConf vectors, which are multi-prototype embeddings based on WordNet knowledge \citep{D16-1174}. 
Similar to the graded Usim experiments, 
adding substitute-based features to embedding features slightly improves the accuracy of the model. Also, combining the CoInCo and WiC data for training does not have a clear impact on results, even in this  binary classification setting.

\section{Discussion}

Results reported for Usim are the average correlation for each target word, but the strength of the correlation varies greatly for different words for all models and settings.
For example, in the case of direct Usim prediction with
embeddings using BERT target, Spearman's $\rho$ ranges from 0.805 (for the verb \textit{fire}) to -0.111 (for the verb \textit{suffer}). 
This variation in performance is not surprising, since annotators themselves found some lemmas harder to annotate than others, as reflected in the Usim inter-annotator agreement measure (Uiaa) \citep{J16-2003}. We find that BERT target word embeddings results correlate with Uiaa per target word ($\rho=0.59, p<0.05$), showing that the performance of this model depends to a certain extent on the ease of annotation for each lemma.
Uiaa also correlates with the standard deviation of average Usim scores by target word ($\rho=0.66, p<0.001$). Indeed, average Usim values for the word \textit{suffer} do not exhibit high variance as they only range from 3.6 to 4.9. Within a smaller range of scores, a strong correlation is harder to obtain.
The negative correlation between Uiaa and Umid ($-0.46, p<0.001$) also suggests that words with higher disagreement tend to exhibit a higher proportion of mid-range judgments.
We believe that this analysis highlights the difference between usage similarity across target words and encourages a by-lemma approach where the specificities of each lemma are taken into account. 

\section{Conclusion}

We applied a wide range of existing word and context  representations to graded and  binary usage similarity prediction. We also proposed novel supervised models which use as features the best performing embedding representations, and make high quality predictions especially in the binary setting, outperforming previous approaches. The supervised models include features based on in-context lexical substitutes. We show that automatic substitutions constitute an alternative to manual annotation when combined with the embedding-based features. Nevertheless, if there is no specific reason for using substitutes for measuring Usim, BERT offers a much more straightforward solution to the Usim prediction problem. 

In future work, we plan to use automatic Usim predictions for estimating word sense partitionability. We believe such knowledge can be useful to determine the appropriate meaning representation for each lemma.

\section{Acknowledgments}

We would like to thank the anonymous reviewers for their helpful feedback on this work. We would also like to thank Jose Camacho-Collados for his help with the WiC experiments. 

The  work  has   been  supported  by  the French National Research Agency under project ANR-16-CE33-0013.

\bibliography{naaclhlt2019}
\bibliographystyle{acl_natbib}

\appendix
\section{Filtering experiments} \label{app:filtering}

Tables \ref{tab:filters} and \ref{tab:lensents} contain results obtained using the different substitute filters described in Section \ref{subusim}. We measure the quality of the substitutes retained in the automatic ranking produced by context2vec after filtering against gold substitute annotations in LexSub data. Here, we only use the portion of LexSub data that does not contain Usim judgments. 

We measure filtered substitute quality against the gold standard using the F1-score, and the proportion of false positives (FP) over all positives (TP+FP). 
Table \ref{tab:filters} shows results for annotations assigned by context2vec using the 
the LexSub/CoInCo pool of substitutes ({\sc auto-lscnc}). Table \ref{tab:lensents} shows results for context2vec annotations with the PPDB pool of substitutes ({\sc auto-ppdb}). 
 
\begin{table}[h] 
\centering
\begin{tabular}{llc} 
Filter & F1 & $FP/(TP+FP)$ \\
\hline
Highest 10 & 0.332 & 0.776\\
Highest 5 & 0.375 & 0.695 \\
PPDB &  0.333& 0.643 \\
GloVe ($T=0.1$) & 0.371 &  0.675 \\
GloVe ($T=0.2$) & 0.373 & 0.661 \\
GloVe ($T=0.3$) & 0.353 & 0.641 \\
c2v score & 0.326 & 0.671 \\
No filter & 0.248 & 0.848 \\

\end{tabular} 
\caption{Results of different substitute filtering strategies applied to annotations assigned by context2vec when using the LexSub/CoInCo pool of substitutes ({\sc auto-lscnc}).} 
\label{tab:filters}
\end{table}

\begin{table}[h] 
\centering
\begin{tabular}{llc} 
Filter & F1 & $FP/(TP+FP)$ \\
\hline
Highest 10 & 0.245 & 0.838\\
Highest 5 & 0.290 & 0.766 \\
PPDB &  0.268& 0.731 \\
GloVe ($T=0.1$) & 0.266 &  0.778 \\
GloVe ($T=0.2$) & 0.268 & 0.769 \\
GloVe ($T=0.3$) & 0.266 & 0.750 \\
c2v score & 0.250 & 0.675 \\
No filter & 0.142 & 0.920  \\

\end{tabular} 
\caption{Results of different substitute filtering strategies applied to annotations assigned by context2vec when using the PPDB pool of substitutes ({\sc auto-ppdb}).} \label{tab:lensents}
\end{table}

\section{Direct Usage Similarity Estimation} \label{app:direct}

Correlations between gold Usim scores for all words and cosine similarities of different embedding types can be found in Tables \ref{tab:embed_correlations1} and  \ref{tab:embed_correlations2}.

\begin{table}[h] 
\centering
\begin{tabular}{ll|c} 
  & Embeddings & Correlation \\
\hline
  \multirow{8}{*}{\parbox{2.2cm}{Full sentence embedding}} & GloVe & 0.142   \\
   & SIF & 0.274  \\
  & c2v  & 0.290 \\
   & USE & 0.272  \\
   & doc2vec & 0.124 \\
   & ELMo av & 0.254 \\
   & ELMo top & 0.248 \\
   & BERT av 4 & 0.289\\
   \hline
   \multirow{6}{*}{\parbox{2.2cm}{Target word embedding}}& ELMo av  & 0.166\\
   & ELMo top & 0.177 \\
   & BERT top & 0.514 \\
   & BERT av 4  & 0.518 \\
   & BERT concat 4 & 0.516 \\
   & BERT 2nd-to-last & 0.486 \\
   \hline
 
   \end{tabular} 
\caption{Correlations of  sentence and word embeddings  on the Usim dataset using different context window sizes (cw). For BERT and ELMo, \textit{top} refers to the top layer, and \textit{av} refers to the average of layers (3 for ELMo, and the last 4 for BERT). \textit{concat 4} refers to the concatenation of the last 4 layers of BERT.} \label{tab:embed_correlations1}
\end{table}

\begin{table}[t] 
\centering
\begin{tabular}{ll|c} 
 Context & Embeddings & Correlation \\
\hline
   \multirow{4}{*}{cw=2} & ELMo top & 0.289\\
   & ELMo av & 0.280\\
   & BERT av 4 & 0.344\\
   & GloVe & 0.140\\
   \hline
   \multirow{4}{*}{cw=3} & ELMo top & 0.282\\
   & ELMo av & 0.279\\
   & BERT av 4  & 0.339\\
   & GloVe & 0.163\\
   \hline
   \multirow{4}{*}{cw=4} & ELMo top & 0.270\\
   & ELMo av & 0.263\\
   & BERT av 4 & 0.311\\
   & GloVe & 0.160\\
   \hline
   \multirow{4}{*}{cw=5} & ELMo top & 0.266\\
   & ELMo av & 0.263\\
   & BERT av 4  & 0.309\\
   & GloVe & 0.162\\
   \hline
  
   \multirow{5}{*}{cw=2 (incl. target)} & ELMo av & 0.284 \\
   & ELMo top & 0.278 \\
   & BERT av 4 & 0.416 \\
   & GloVe & 0.159  \\
   & USE & 0.146\\
   \hline
   \multirow{5}{*}{cw=3 (incl. target)} & ELMo av & 0.280 \\
   & ELMo top & 0.273 \\
   & BERT av 4 & 0.395 \\
   & GloVe & 0.180  \\
   & USE & 0.184\\
   \hline
   \multirow{5}{*}{cw=4 (incl. target)}  & ELMo av & 0.267 \\
   & ELMo top & 0.265 \\
   & BERT av 4 & 0.365 \\
   & GloVe & 0.176  \\
   & USE & 0.191\\
   \hline
   \multirow{5}{*}{cw=5 (incl. target)} & ELMo av & 0.266 \\
   & ELMo top & 0.263 \\
   & BERT av 4 & 0.359 \\
   & GloVe & 0.175  \\
   & USE & 0.221\\
   \hline
  
      \end{tabular} 
\caption{Correlations of different sentence and word embeddings on the Usim dataset using different context window sizes (cw).} \label{tab:embed_correlations2}
\end{table}

\section{Feature Ablation on Usim} \label{app:ablation}

Results of  feature ablation experiments on the Usim development sets are given in Table \ref{tab:ablation}.

\begin{table*}[t!]
\centering

\begin{tabular}{l|cccccc} 
Ablation & Gold & {\sc auto-lscnc} & {\sc auto-ppdb} \\
\hline
None & 0.729 & 0.538 & 0.524 \\
Sub. similarity & 0.701 & 0.537 & 0.524 \\
 Common sub. & 0.722& 0.538 & 0.524 \\
  GAP & 0.730& 0.537& 0.523 \\
   c2v & 0.730& 0.539 & 0.523\\
    Bert av 4 target & 0.700 &  0.348 & 0.283\\

\end{tabular}

\caption{Results of feature ablation experiments for systems trained and tested on the Usim dataset with gold substitutes (\textit{Gold}) as well as automatic substitutes from different pools, Lexsub/CoInCo ({\sc auto-lscnc}) and PPDB ({\sc auto-ppdb}). Rows indicate the feature that is removed each time. Numbers correspond to the average Spearman $\rho$ correlation on the development set across target words.} \label{tab:ablation}

\end{table*}

\section{Dev experiments on WiC} \label{app:devwic}

Table \ref{tab:devwic} shows the accuracy of different configurations on the WiC development set.

\begin{table*}[ht!]
\centering

\begin{tabular}{ l l |  c}
  Training set & Features  &  Accuracy\\ 
  \hline
  \multirow{8}{*}{WiC}  & BERT av 4 last target word  & 65.24 \\
   & c2v  & 57.69 \\
   & ELMo top cw=2  & 61.11 \\
   & USE  & 63.68 \\
   & SIF  & 60.97 \\
   & Only substitutes  & 55.41 \\
   & BERT av 4 target word \& USE   & \textbf{67.95} \\
   & Combined  & \textbf{66.81} \\
  
   \hline
   \multirow{8}{*}{WiC + CoInCo}   
   & BERT av 4 target word  & 64.96 \\
   & c2v  & 58.12 \\
   & ELMo top cw=2  & 61.11 \\
   & USE  & 63.53 \\
   & SIF  & 59.97 \\
   & Only substitutes  & 56.13  \\
   & BERT av 4  target word \& USE  & \textbf{68.66}\\
   & Combined  & \textbf{66.81} \\

\end{tabular}

\caption{Accuracy of different features and combinations on the WiC development set. On this dataset, the two best types of embeddings, that were chosen for the Embedding-based and Combined configurations, were BERT (target word, average of the last 4 layers) and USE. Both Only-substitutes and Combined use features of automatic substitutes from the PPDB pool, and back off to the Embedding-based model when there were no paraphrases available for the target word in the PPDB.}
\label{tab:devwic}
\end{table*}

\end{document}